\documentclass[journal,onecolumn]{IEEEtran}
\usepackage{graphicx}
\usepackage{tikz}
\usepackage{comment}
\usepackage{amsmath,amssymb} 
\usepackage{color}
\usepackage{wrapfig}
\usepackage{xspace}
\usepackage[capitalize]{cleveref}
\usepackage{booktabs}
\usepackage{url}
\usepackage{wrapfig}
\usepackage{caption}
\usepackage[accsupp]{axessibility}  
\newcommand*{\eg}{e.g.\@\xspace}
\newcommand*{\ie}{i.e.\@\xspace}
\newcommand{\etal}{\textit{et al.~}}

\begin{document}
\pdfcompresslevel=9
\pagestyle{headings}
\title{Intuitive Shape Editing in Latent Space}
\author{Tim Elsner*\thanks{*elsner@cs.rwth-aaachen.de}, Moritz Ibing, Victor Czech, Julius Nehring-Wirxel, Leif Kobbelt\\RWTH Aachen University}

\maketitle

\begin{abstract}
The use of autoencoders for shape editing or generation through latent space manipulation suffers from unpredictable changes in the output shape. Our autoencoder-based method enables intuitive shape editing in latent space by disentangling latent sub-spaces into style variables and control points on the surface that can be manipulated independently. The key idea is adding a Lipschitz-type constraint to the loss function, i.e.\ bounding the change of the output shape proportionally to the change in latent space, leading to interpretable latent space representations. The control points on the surface that are part of the latent code of an object can then be freely moved, allowing for intuitive shape editing directly in latent space. We evaluate our method by comparing to state-of-the-art data-driven shape editing methods. We further demonstrate the expressiveness of our learned latent space by leveraging it for unsupervised part segmentation.
\end{abstract}

\section{Introduction}
\begin{wrapfigure}[24]{r}{0.55\textwidth}
  \includegraphics[width=\linewidth]{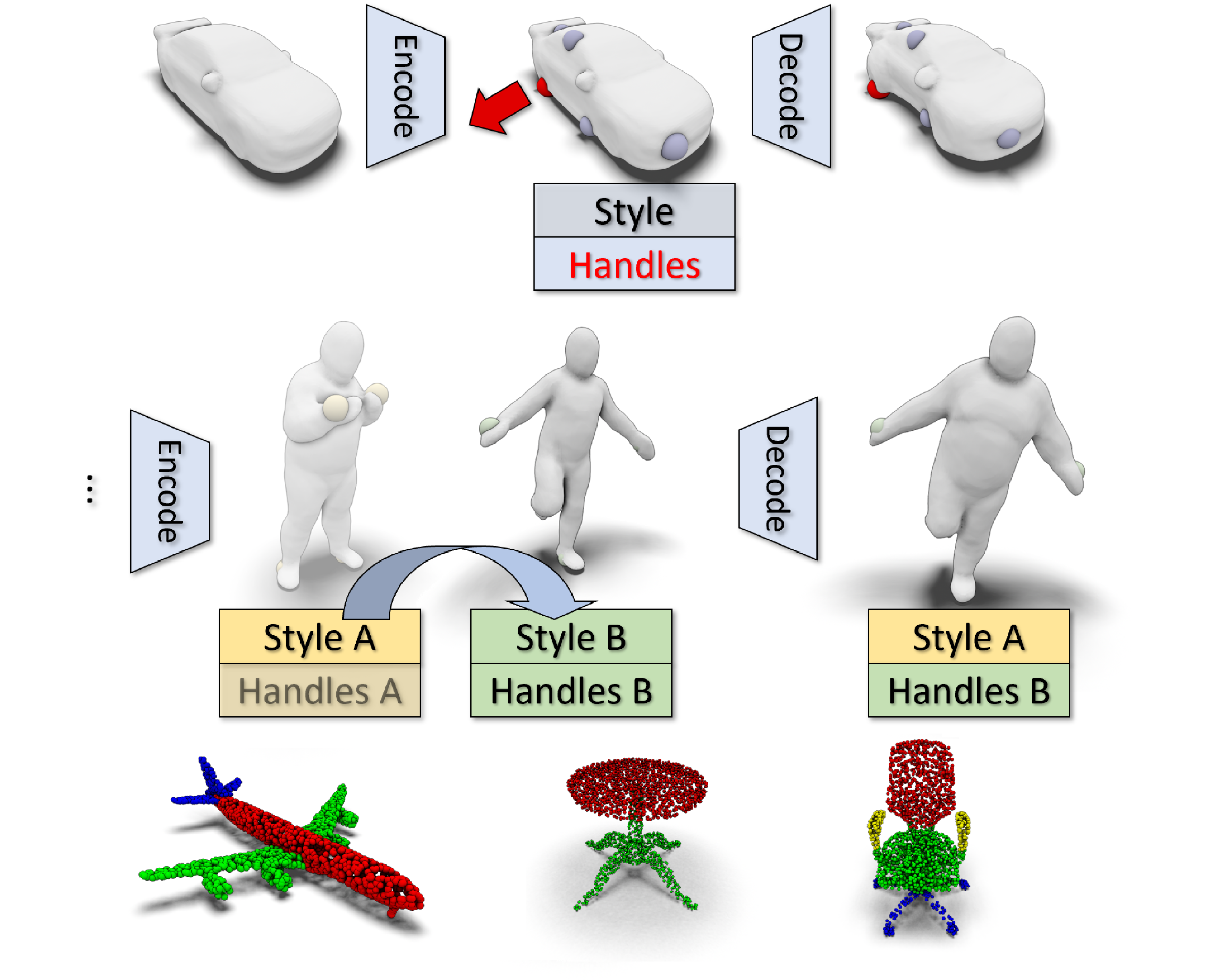}
  \caption{Proposed applications of our learned latent space: Shape editing, transferring the style from one humanoid to the pose of another, and part segmentation.}
   \label{fig:fig1}
\end{wrapfigure}

Creating new 3D shapes and editing existing ones is a key task in computer graphics. Having the ability to interactively design shapes is an ongoing part of research \cite{sachs19913}\cite{hagenlocker1998cffd}\cite{liu2021deepmetahandles}\cite{dekkers2009sketching}. In particular, designing different variations of an item from a collection of shapes can be cumbersome. These applications can benefit from tools that guide the user with learned semantic expertise, such as enforcing symmetry for airplane wings, or having chair legs attached to the seating area.
Approaches in deep learning like Generative Adversarial Networks (GANs)\cite{goodfellow2014generative} can be used to generate new examples of a given shape
collection\cite{li2018point}, but usually offer little intuitive manipulability\cite{karras2019style}\cite{karras2020analyzing}, or require complex regularisation techniques \cite{DBLP:journals/corr/ChenDHSSA16}\cite{DBLP:journals/corr/MirzaO14} that can not be transferred to geometry easily. 

Autoencoders\cite{kramer1991nonlinear} are an established way to learn efficient encodings without any supervision and provide a parametric representation of a given shape collection that is in principle manipulable. While a basic autoencoder, and even more so its derivatives\cite{kingma2013auto}\cite{razavi2019generating}\cite{higgins2016beta}, can be used for shape editing, this process is not controllable or intuitively interpretable.
Motivated by the observation that the most user-friendly forms of shape editing tools are built around controlling the surface through manipulating control points (\textit{handle points}) on the shape directly\cite{herndon1994challenges}, we propose an autoencoder that does not use some arbitrary, hard to interpret latent code, but uses a small number of interpretable 3D handle points instead. The handle points themselves represent the coarse structure of the shape. For surface details or regions unrelated to handle point, we introduce a style vector holding all information that is not fully determined by the handle points. We disentangle style information and handle information by enforcing any latent change to be proportional to the output change. As we need consistent handle points across shapes for this, we also provide a simple technique to obtain these for a shape collection.\\
Different options can be considered to represent geometry in an autoencoder: Explicit representations like point clouds can suffer from artifacts. For example, having the same chair with a wider backrest will either require more points or to spread out the existing ones, resulting in inconsistent point density. Voxelisation requires large amounts of memory to work with reasonable resolutions. Generating and processing triangle meshes is complex and cumbersome, as convolutions are complicated to define and properties like 2-manifoldness are difficult to produce\cite{feng2019meshnet}\cite{Wang_2018_ECCV}. Hence, we propose to use an implicit autoencoder, based upon the implicit decoder DeepSDF by Park \etal \cite{park2019deepsdf}.\\
\paragraph*{Contribution}
This paper introduces a new way of obtaining interpretable latent code consisting of manipulable handle points and an interchangeable style vector in an autoencoder setup. This allows for interactive adaptive shape editing as shown in \cref{fig:fig1}. The coarse structure of the shape can be controlled through the handles directly, while the style can be changed independently. To obtain proper disentanglement of handles and style and to assure a smooth transition when adjusting the handles, we introduce a regularisation technique that tethers change in the latent space representation to a respective change in the output. This means that small movements to the handles or the style cause small differences in the output, and that applying changes to the style and handles is independent of each other.
We further introduce a technique allowing changes in one part of the shape to adaptively propagate, \eg keeping airplane's wings symmetric. To demonstrate the semantic power of the learned latent space, we show a simple part segmentation approach. We actively avoid complicated architectures or data augmentation to show the versatility and strength of the approach as basis for future work. We further compare to other shape editing approaches.

\section{Related Work}
To make our latent space editable and allow shape deformations, we rely on disentangling the latent space into 3D handle points in shape-space and a style vector. Hence, our work is mostly related to approaches either applying shape deformations or disentangling the latent space representation of an input.
\paragraph*{Disentanglement}
Disentangling an input into semantically different components is an ongoing part of research. Approaches like (Beta-)VAEs\cite{higgins2016beta} offer such disentanglement, which can also be transferred to geometry \cite{xiaopoint}\cite{tan2018variational}. This yields decomposition of the latent space in different attributes, but lacks interpretability. Fumero \etal \cite{DBLP:journals/corr/abs-2103-01638} disentangle by enforcing that changing one factor of the input only changes one latent code.
Aumentado-Armstrong \etal \cite{aumentado2019geometric} split the latent space into a part representing the intrinsic geometry of the shape and its extrinsic pose. However, their latent space is still not interpretable, while we explicitly describe the pose of our shape through handle positions. Methods like \cite{yang2020dsm} or \cite{mo2019structurenet} disentangle shapes into a structure part represented by a part hierarchy and shape details encoded in latent vectors. However, their output geometries are only controllable through (unintuitive) latent manipulation. To ease disentanglement, some approaches like Yang \etal \cite{yang2020dsm} or Zhou \etal\cite{zhou2020unsupervised} utilize (partly) labeled input data. Lastly, approaches using conditional GANs \cite{chen2021decor}\cite{abdal2021styleflow} also shown promising results.
\paragraph*{Shape deformation}
There are many previous works that discuss how one shape can be deformed to match the geometry of a target shape as close as possible. By restricting the deformation space, a shape still keeps the topology and some details of the original shape, but the possible alignment is rather limited. The presented methods differ mostly in the way they represent the deformation space. Yumer \etal \cite{yumer2016learning} compute control points in a grid and use those to describe transformations, while Shechter \etal \cite{shechter2022deepmls} use control points on the surface. Gadelha \etal \cite{gadelha2020learning} generate shapes handles that can be used for deformation, while Gao \etal \cite{gao2019sdm} learn part-based deformations and Uy \etal\cite{uy2021joint} learn a shape space through them. Controlling the shape space by keeping it metric can also be used to \cite{DBLP:journals/corr/abs-2003-12283} and is similar to our Lipschitz-inspired regularisation term.
Other approaches deform from a source to a target shape by deforming a geometric proxy \cite{hanocka2018alignet}\cite{yifan2020neural}, allowing deformations from source to partial targets by learning grid-based deformations. Lui \etal \cite{liu2021deepmetahandles} propose grouping handles working with biharmonic coordinates together into metahandles that can be manipulated. The resulting disentangled handles smoothly edit the shape, but are not part of the shape space itself. Sung \etal \cite{sung2020deformsyncnet} learn transfer deformations from one shape to another by parameterizing the deformations themselves through an autoencoder.  Atzmon \etal \cite{atzmon2021augmenting} on the other hand do not try to give an intuitive transformation operator, but restrict latent interpolation to adhere to deformation constraints. In contrast, our approach does not transform a source shape or uses proxies, but instead the shape itself is parameterized (in part) by the handles.

\begin{figure*}[t]
  \centering
    \includegraphics[width=0.9\linewidth]{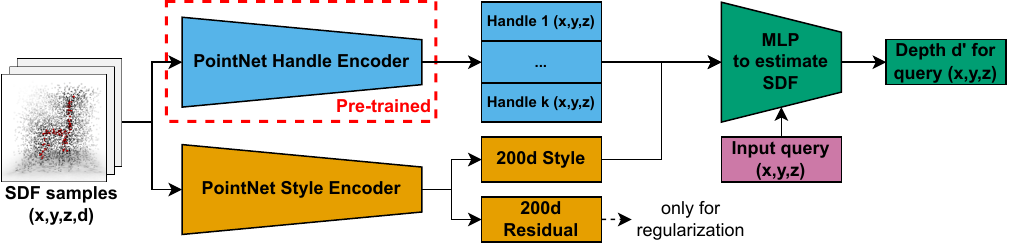}
    \caption{Our proposed architecture: Embedding and encoding an SDF through a simplified PointNet to obtain handles (blue, $f_{H}$), style and residual (orange, $f_{S}$ and $f_{R}$), then reconstructing the distance values from an embedding and a coordinate through an MLP. The handle encoder is pre-trained, then frozen.}\label{fig:architecture}
\end{figure*}
\section{Learning Adaptive Shape Editing}
Our network is trained unsupervised as an autoencoder on a shape collection. As input, it takes an object sampled as SDF, produces a latent code composed of actual 3D points on the surface (\textit{handles}) and additional information (\textit{style}), and then decodes this latent code together with a point to obtain a distance value. After training, the latent code of an object can be used for manipulations, \eg the handles can be moved to perform shape editing.\\
This is achieved through using the following network architecture (see \cref{fig:architecture} for high-level and Appendix \ref{sec:details} for details): The encoder is computing three distinct pieces: First, a set of 3D handles on the surface of the shape holding most of the coarse structural information. Second, a style vector that contains any other information of the shape. The third component is the residual, which is only utilized during training to regularize the other two components and acts as a
buffer to avoid inconsistent style vectors. The decoder then predicts the depth values for the given input sample positions from the handles and style vector.

After training, we can encode a shape, move one or multiple of the handles or replace the style vector (\textit{style transfer}), and decode the result. However, applying certain operations like moving only one side of the wings of an aircraft, are not \textit{plausible} in the sense of the shape collection: These latent space configuration do not lie on the manifold of learned shapes that our decoder can properly decode. To find the most similar, nearest plausible shape on the manifold, we apply changes such as handle position shifts gradually throughout iterations of de- and encoding.
\paragraph*{Difference between handles and style}
We first need to define how the handles should control the output, and define style implicitly as the remaining details: Given two shapes $A(H_A)$, $B(H_B)$, with handle positions $H_A$, $H_B$, style refers to the difference between $A(H_B)$ and $B(H_B)$, \ie the aspects of the shape not defined smoothly by handle position alone. For example, for two chairs with almost identical handle positions $H_A \approx H_B$ that only differ by a hole in the backrest, we consider the difference in shape $A(H_A) - B(H_B)$ a stylistic difference. The definition of style is therefore directly related to the number of handle points. For an example of this, consider the car used in the top-left of Appendix \ref{app:more_results}.
\paragraph*{Desired properties}
The latent code of our autoencoder consists of two components: the handle
component, controlling the global shape or pose, and the style component, encoding the shape details. When training our network we have to make sure that the latent embedding does not degenerate. Potential threats are that the handles can
"over-encode" style by micro-offsets causing style changes rather than shape changes.
Conversely, the style vector could encode global shape information rendering
the handles useless. We prevent the former by introducing a Lipschitz-inspired term
(\cref{eq:lip}) into the loss function that forces changes in the latent code to be of the
same magnitude as the resulting changes in the output shape. We prevent
the latter by a term (\cref{eq:spen}) that penalizes changes in the style component
if the shape difference can be expressed as handle motion. In addition to this, we formalize a loss to keep handles and style as independent and exchangeable as possible (\cref{eq:ind}).

\subsection{Encoder}
For the encoder, we first learn to encode the desired handle positions for each shape, then freeze the part of the encoder that produces the handles. We then train the encoder to produce residual and style, guided by our Lipschitz-inspired regularization term to encode what is not already expressed smoothly by the handle positions. Note that the residual acts only as a buffer for better generalisation. 
\\
In detail, the encoder $f$ of our network consists of three distinct parts $f_{H}$, $f_{S}$, and $f_{R}$ that take a uniform random sampling of a shape's SDF, $S_u$, as input. $S_u$ is a set of points $p$ with position $p_{xyz}$ and distance $p_d$. All three encoders are based on a simplified PointNet\cite{qi2017pointnet} architecture that embeds a 4D point cloud sampled from an SDF, \ie sample position and distance. The embedding vector is then processed through a multilayer perceptron (MLP). For technical details, see \cref{fig:architecture} and Appendix \ref{sec:details}.
We require consistent handles across shapes either calculated from a method such as given in \cref{sec:canonical_handles}, or given by the user. The handle encoder $f_{H}(S_u)$ is pre-trained to produce the given handle position. For this, the embedding vector is encoded through an MLP to form our desired handles, using a simple L2 loss. We then freeze the learned encoder part, not allowing change to the learned handles while training the rest of the network, consisting of style, residual, and decoder.\\
Learning the style encoder $f_{S}(S_u)$ and the residual encoder $f_{R}(S_u)$, we don't train for specific latent values, but instead add regularization: To force the style vector to be used in a meaningful way and not to over-encode the information in the subtle changes of the handles, we are tethering the change in shape space to the change in latent space. We formulate a Lipschitz-inspired constraint per trained batch that enforces the change in the combined latent variable of style, residual, and handles to be proportional to the change in the output. In consequence, the occurring difference in the SDF of two shapes that can not be explained by the handles alone then leads to the style variable to explain all the details that differ. We thus define our loss $L_{d}$ regarding a batch of items $B$ and all successive pairs of items $i, i+1$ (to avoid a quadratically increasing number of comparisons with increasing batch sizes): 

\begin{equation}
  \begin{split} 
  L_{d} = \sum_{i} (D_{input}(i, i+1) - D_{output}(i, i+1))^2
  \end{split}
  \label{eq:lip}
\end{equation}

Where $D_{input}$ defines the pairwise difference between two sampled shapes, given as their weighted distance differences to the surface for the same sample points used for all shapes in the batch, \ie:
\begin{equation}
  \begin{split} 
  D_{input}(a,b) = \frac{\sum_{p \in S_u^a, q \in S_u^{b}} |p_d - q_d|(w(p_d, S_u^{a}) + w(q_d, S_u^{b}))}{2|S_u^a|}
  \end{split}
\end{equation}
To improve quality where it matters most, \ie close to the surface, the weight of a point with distance $d$ for a sampling $S$ in the loss term is weighted by the distance to the surface: $w(d,S) = \max_{p_d \in S} |p_d| - |d|$.\\
For $D_{output}$, we simply add the L2 norm between the style vectors, the residual vectors, and each handle point pair. We further add an individually learned weight to each handle. This is necessary because every region and the handle controlling it can have a different impact on the change of the SDF (\eg consider change in the thin legs of a chair versus change in its seating area). Recall that while we learn the weights of each handle, \ie their influence on the change of the shape, the handle positions themselves are kept fixed.\\
In summary, $L_{d}$ penalizes the imbalance between the shape difference and the latent distance for pairs of shapes. By using $L_{d}$, we obtain a latent space where change applied in latent is proportional to the change observed in the SDF, a key attribute for the desired editability of our latent space.

\paragraph*{Keeping handles relevant}
To avoid disregarding handles and just encoding everything in style, we introduce an additional penalty on the style and the residual. We define:
\begin{equation}
  L_{spen} = ||f_{S}(S_u)||_2, ~~~~
  L_{rpen} = ||f_{R}(S_u)||_2
  \label{eq:spen}
\end{equation}
$L_{d}$ encourages two items with identical handles having all their differences encoded in their style vector; Opposing to that, $L_{spen}$ encourages that for two items with very different handles, the difference in their handles is actually used in the reconstruction process as much as possible and their different looks are not just encoded in their style vectors, ignoring the handles.
To capture differences between style and handle changes that can not be properly reflected in the loss and would otherwise conflict with $L_{d}$, the residual part of the network is used: For two identical chairs where one of them is scaled to be wider, the resulting difference in the SDF can be explained according to $L_{d}$ by the change in handles. As this does not always work in a smooth and linear fashion, the resulting style vectors could be encouraged to differ. The residual absorbs these differences in a way that does not harm the generalisation of the style vector.
In summary, the encoder produces a latent code consisting of style, residual, and handles that is regularized to be changing smoothly and proportional to change in the shape. This means having very similar input shapes produce very similar latent codes, and preferably not to encode coarse structural or pose information into the style vector. This results in a latent space disentangled into style and handle positions.
\subsection{Decoder}\label{sec:full}
We now describe the decoding process $g(f_{H}(S), f_{S}(S), p_{xyz})$. The latent code, consisting of the encoded style and handles, is concatenated to a 3D point $p_{xyz}$. We then apply a MLP to predict the distance value $p_d$. The latent code produced by the residual encoder $f_{R}(S_u)$ is not used for the decoder.
Now we define a reconstruction loss that measures the L2-difference between input distance and predicted signed distance for each point. To increase the quality of the learned SDF near the surface, we use another improvement inspired by Park \etal \cite{park2019deepsdf} to make the SDF sharper at the contour of the shape. We decode an additional set of Gaussian samples $S_g$ around the surface of the shape. This increases quality where it is most visible. Hence, we obtain a reconstruction loss that is composed of two parts (uniform and close-to-surface points) where the average weight over all samples sums up to one: 

\begin{equation}
  \begin{split}
  L_{rec} = &\sum_{p \in S_u} \frac{|g(f_{H}(S_u), f_{S}(S_u), p_{xyz}) - p_d| w(p_d, S_u)}{\sum_{q \in S_u} w(q_d, S_u)}\\
  +&\sum_{p \in S_g} \frac{|g(f_{H}(S_u), f_{S}(S_u), p_{xyz}) - p_d| w(p_d, S_g)}{\sum_{q \in S_g} w(q_d, S_g)}
  \end{split}
  \label{eq:rec}
\end{equation}

\paragraph*{Enforcing independence of style and handles}
To guarantee that our editing process works as intended, we want the style and the handles to be independent of each other: Any combination of style and handles must still be meaningful, and an occurring change in style should not apply change to the handle positions and vice-versa. Again, as with $L_{d}$, we define the loss term batch-wise by splicing each element in the batch $i$ with the next element $i+1$:
\begin{equation}
  \begin{split}
  &L_{ind} =\\
  &||f_{H}(g'(f_{H}(S_u^{i}), f_{S}(S_u^{i+1}), (S_u^{i})_{xyz})) - f_{H}(S_u^{i})||_2\\
  +&||f_{S}(g'(f_{H}(S_u^{i+1}), f_{S}(S_u^{i}), (S_u^{i})_{xyz})) - f_{S}(S_u^{i})||_2
  \end{split}
  \label{eq:ind}
\end{equation}
where $g'$ means applying $g$ on a set of points while $g'$ also provides $xyz$ coordinates of each point as output to ease notation (instead of concatenating the sample $xyz$ with the computed distance). For encoding a shape, adapting its handle (style) vector, and then de- and re-encoding the shape again should not cause any changes in the style (handles).
For the complete loss function, we now obtain: 
$$L = \lambda_{1}L_{rec} + \lambda_{2}L_{d} + \lambda_{3}L_{ind} + \lambda_{4}L_{rpen} + \lambda_{5}L_{spen}$$
With $\lambda_{1}=\lambda_{2}=\lambda_{3}=100$, $\lambda_{4}=1.0$ and $\lambda_{5}=0.1$.
We learn to describe an input shape through a set of handle points that must be used in the reconstruction process guided by $L_{rec}$. Any differences between shapes not encoded smoothly in the handles themselves are instead encoded in the style vector by $L_{d}$. Style and handles are kept independent and exchangeable by $L_{ind}$, while $L_{spen}$ prevents ignoring the handles, and $L_{rpen}$ forces the residual to only be used if necessary.
\subsection{Canonical Handles}\label{sec:canonical_handles}
To have handles consistently at corresponding locations for shapes of a collection that we can use for editing, we need a mapping across different shapes: For intuitive editability purposes, a handle point that lies on the tip of the left wing of one airplane should be in a similar spot for another airplane. We call these consistently mapped handle points \textit{canonical handles}. A range of approaches exist to find correspondences between shapes\cite{schmidt2020inter}\cite{ezuz2019reversible}\cite{schreiner2004inter}, but these are often slow, require well behaved geometry, or additional supervision.
Instead, we use a simple procedure to find consistent correspondences between shapes of a collection. This is similar to what \cite{li2021sp} leverage for their correspondences:
To obtain a canonical set of handles that roughly correspond to the same part on every shape, we train a point cloud autoencoder called \textit{canonicalizer}. We use PointNet as encoder and an MLP as decoder, and minimize Chamfer distance \cite{fan2017point} between input and output. The output neurons then localize, where the same index of the output vectors is a part of the same component throughout all shapes in the collection. These correspondences can then be used to obtain consistent, canonical handle positions throughout a shape collection. We compute a mean shape of the shape collection from the autoencoded shapes, taking the mean position for every output neuron triplet of the MLP that represents a point. We then apply farthest point sampling to obtain our handles. The exact architecture and the produced correspondences and handles can be seen in Appendix \ref{app:canonicaliser}.

\begin{figure*}[t]
	\hfill
	\begin{minipage}{0.45\columnwidth}
		\centering
        \includegraphics[width=\linewidth]{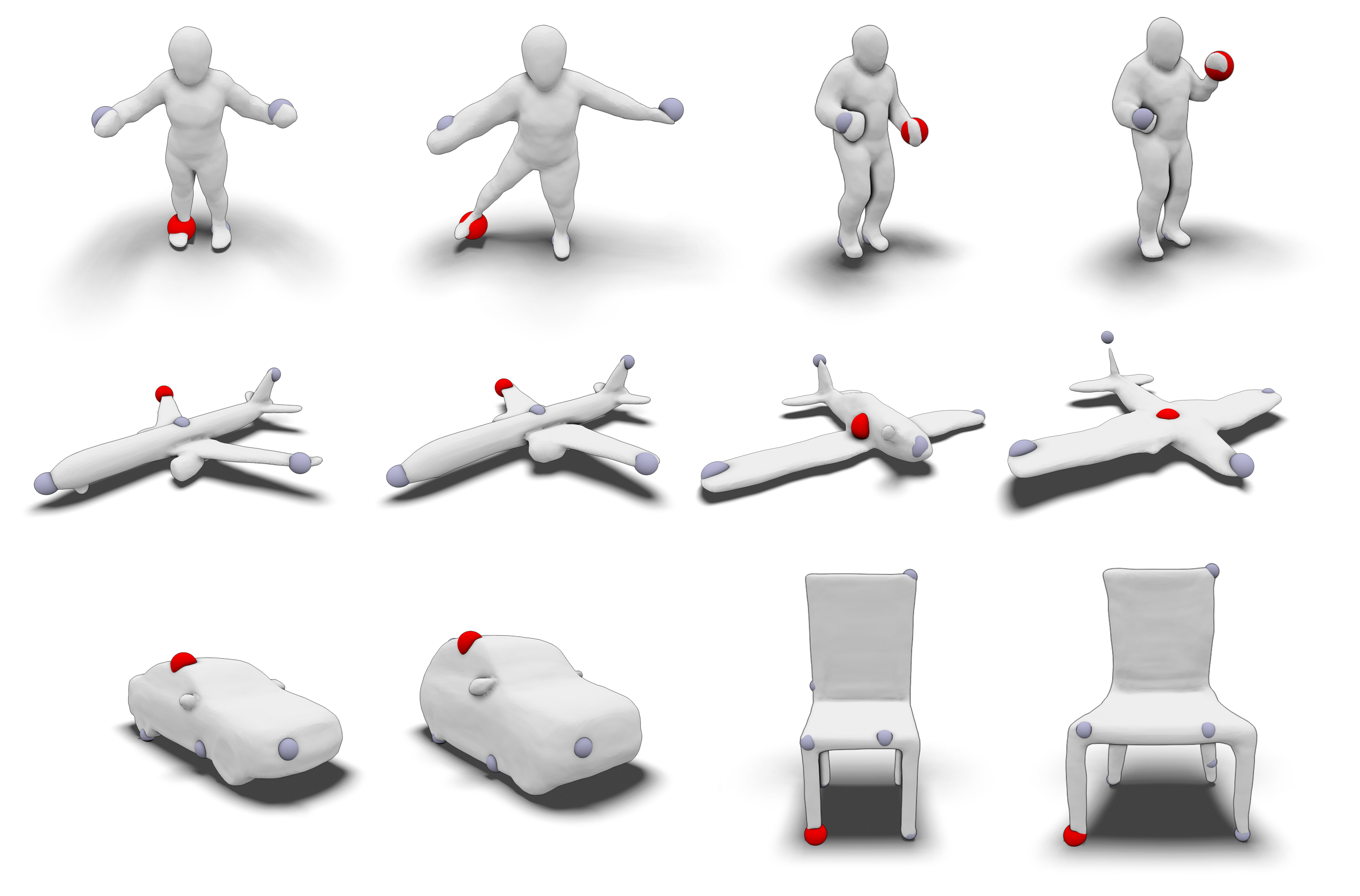}
        \caption{Examples of editing shapes moving the handles (red, always pairs of before- and after editing).}\label{fig:handleediting}
	\end{minipage}
	\hfill
	\begin{minipage}{0.45\columnwidth}
		\centering
		\includegraphics[width=\linewidth]{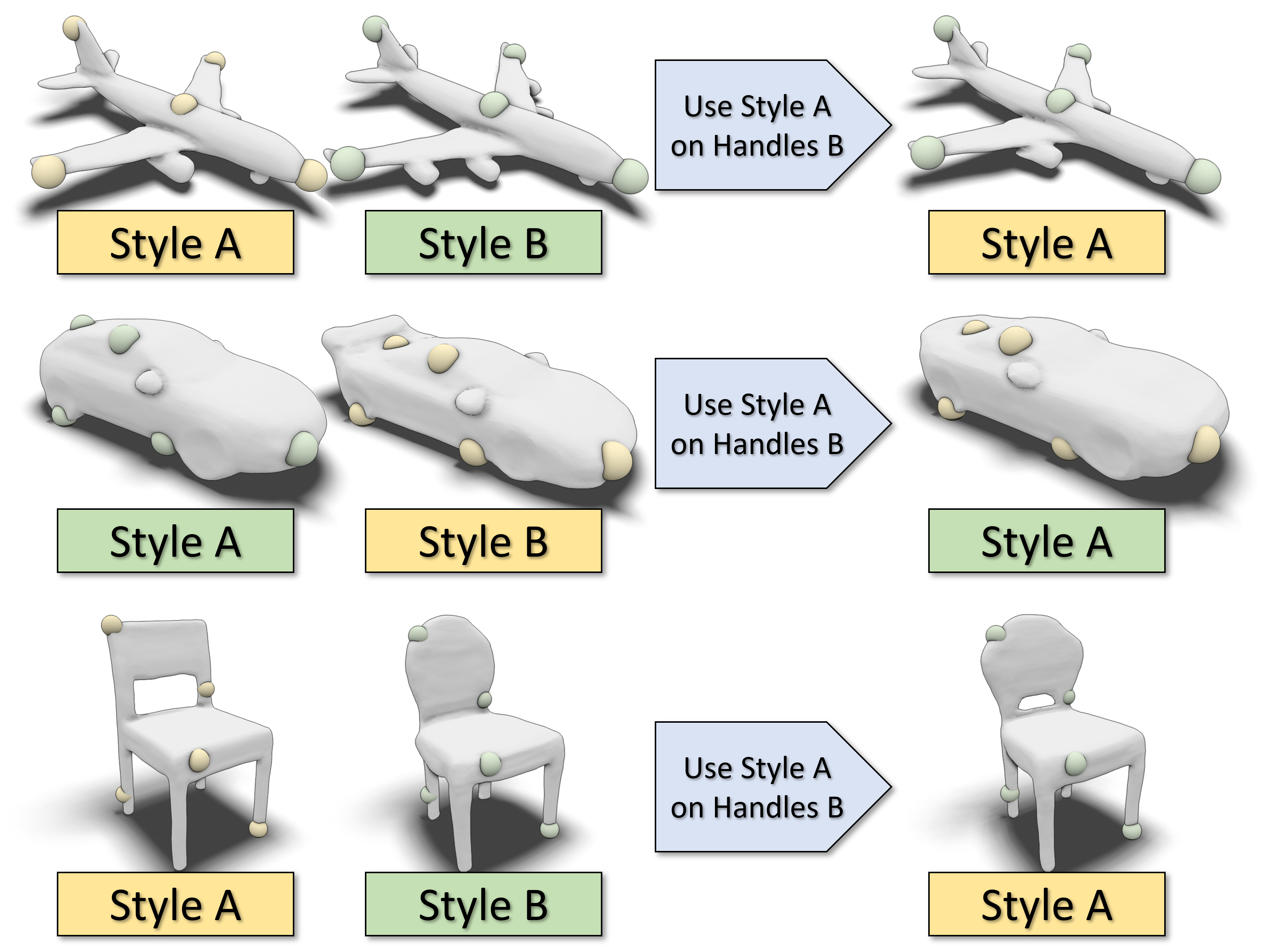}
        \caption{Examples of editing shapes by transferring style through exchanging style vectors.}\label{fig:styletransfer}
	\end{minipage}
	\hfill
\end{figure*}

\subsection{Applying Adaptive Shape Editing}
\begin{wrapfigure}{l}{0.38\textwidth}
    \centering
    \includegraphics[width=0.375\textwidth]{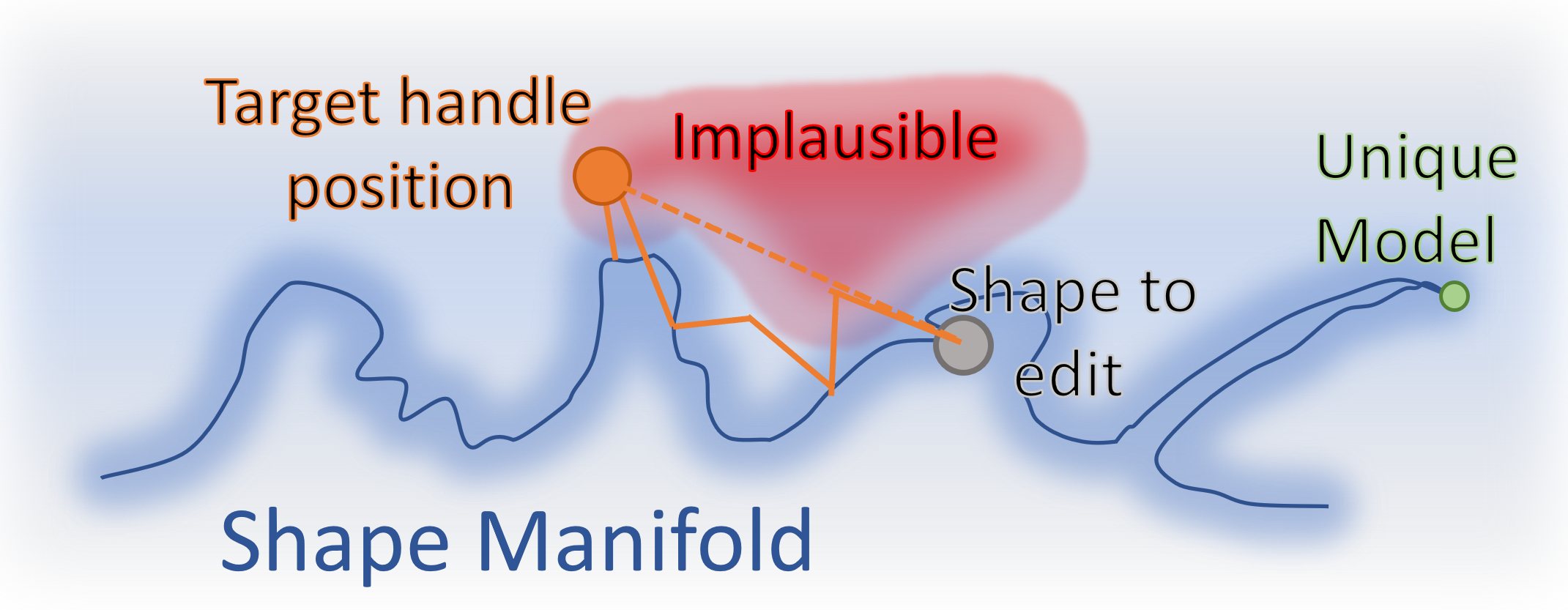}
\end{wrapfigure}
To perform adaptive shape editing, a given shape is encoded. Through \eg a user or an automated script, the editing is performed by changing the style vector and/or the handles and decoding these again. While this process already works for some examples, the applied change does not pass on to the unedited parts completely, \eg not keeping airplane wings symmetric. To achieve this, we leverage a similar idea as denoising autoencoders\cite{vincent2008extracting}, where a noised input is mapped back to the original input. Through autoencoding a gradually modified latent code of a shape, we effectively project back to the latent manifold of plausible shapes, as has been described by Lei \etal \cite{lei2020geometric}. And indeed, we observe that for slightly noised latent codes of examples from our shape collection in the latent code, de- and encoding them again moves them closer to the original latent code, hence back to the manifold of plausible shapes.
We further observe that very unique items form their own sub-part of the manifold, offering fewer plausible latent directions for editing. 
\paragraph*{Editing in latent space and keeping shapes meaningful}\label{sec:manifold_projection}
In practise, only regions of the latent space close to the shape manifold are properly regularized. If we move too far from the manifold, iterating the decoding and encoding process will not move us back to a plausible shape. To avoid this, we apply all edits to the shape through multiple rounds. We gradually apply change to the latent code by moving a handle at most by $0.075$ units towards the target direction for a shape collection normalized to the unit cube. For all shape edits in this paper, we used at most 10 rounds and ended earlier if we had the handle move less than $0.03$ units close to its target in an iteration. This is to prevent iterations where we keep projecting back to the manifold from an implausible configuration, \eg when moving a chair leg beneath the ground plane. For an example of different editing operations using handles, see \cref{fig:handleediting}. We also keep the style vector fixed for handle editing.
To transfer the style from one shape to another, we simply encode two shapes, swap out their style vector, and decode them. For an example of different style transfer examples, see \cref{fig:styletransfer}.

\section{Evaluation}
To evaluate our framework, we picked shape collections with varying characteristics. For our chosen man-made shapes (airplanes, tables, cars, chairs from Shapenet\cite{chang2015shapenet}) and organic shapes (FAUST \cite{bogo2014faust}), semantic properties like learning symmetry, physical properties like no freely floating parts, and plausible poses of bodies have to be learned.\\
We first compare our method to other state-of-the-art methods, then evaluate our qualitative results for handle- and style manipulation. While providing quantitative results for a shape editing framework is somewhat ill-posed, we instead provide quantitative analysis of the key properties that a shape editing tool should have: Expressiveness and the learned underlying semantics used in the editing process. We first provide experiments that measure the latent space properties used for editing. For expressiveness, we then utilize the same metric as \cite{liu2021deepmetahandles}. Regarding semantics, we compare ourselves to part segmentation approaches, using an ad hoc way of extracting a segmentation from our learned network.
\begin{figure*}[t]
	\begin{minipage}{0.3\columnwidth}
		\centering
		\includegraphics[width=\linewidth]{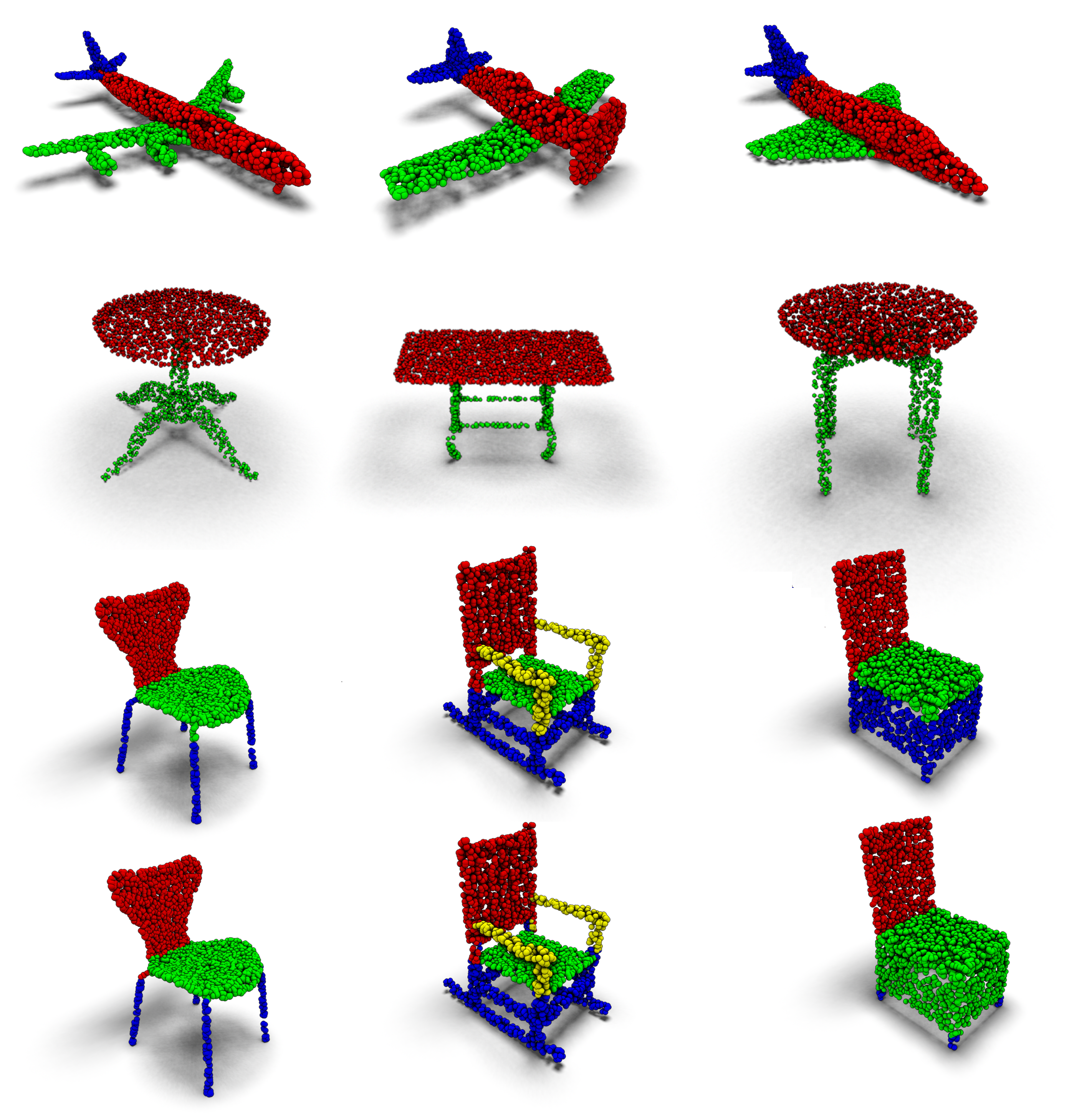}
        \caption{Part segmentation, with the last row showing the ground truth for chairs.}\label{fig:segmentation_results}
	\end{minipage}
	\hfill
	\begin{minipage}{0.375\columnwidth}
		\centering
		\includegraphics[width=\linewidth]{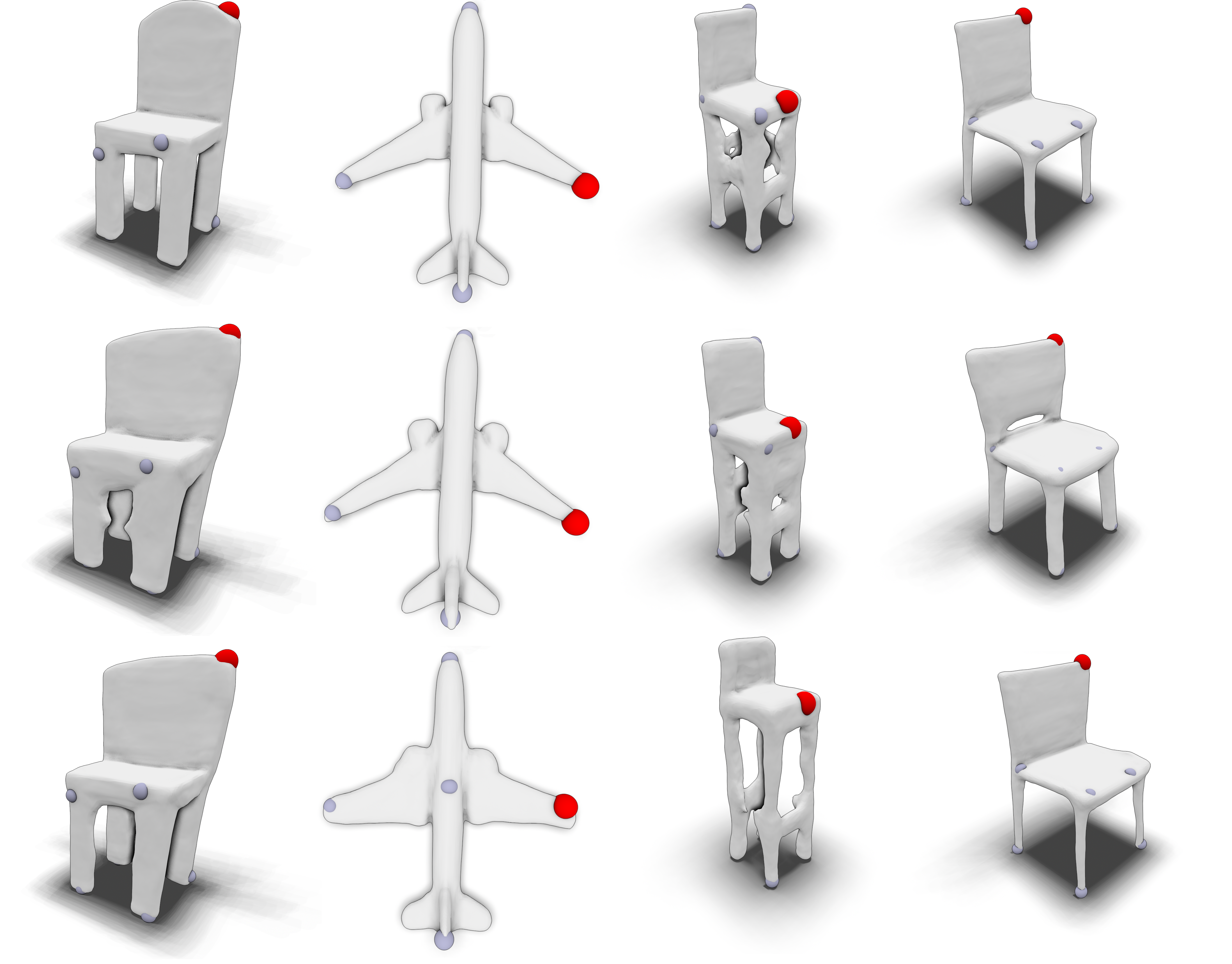}
        \caption{Ablation. Top to bottom: Input, ablated, complete. Left to right: No residual, no independence loss, no style penalty, no Lipschitz-inspired.}\label{fig:ablation}
        \vspace{-10pt}
	\end{minipage}
	\hfill
	\begin{minipage}{0.225\columnwidth}
		\centering
		\includegraphics[width=\linewidth]{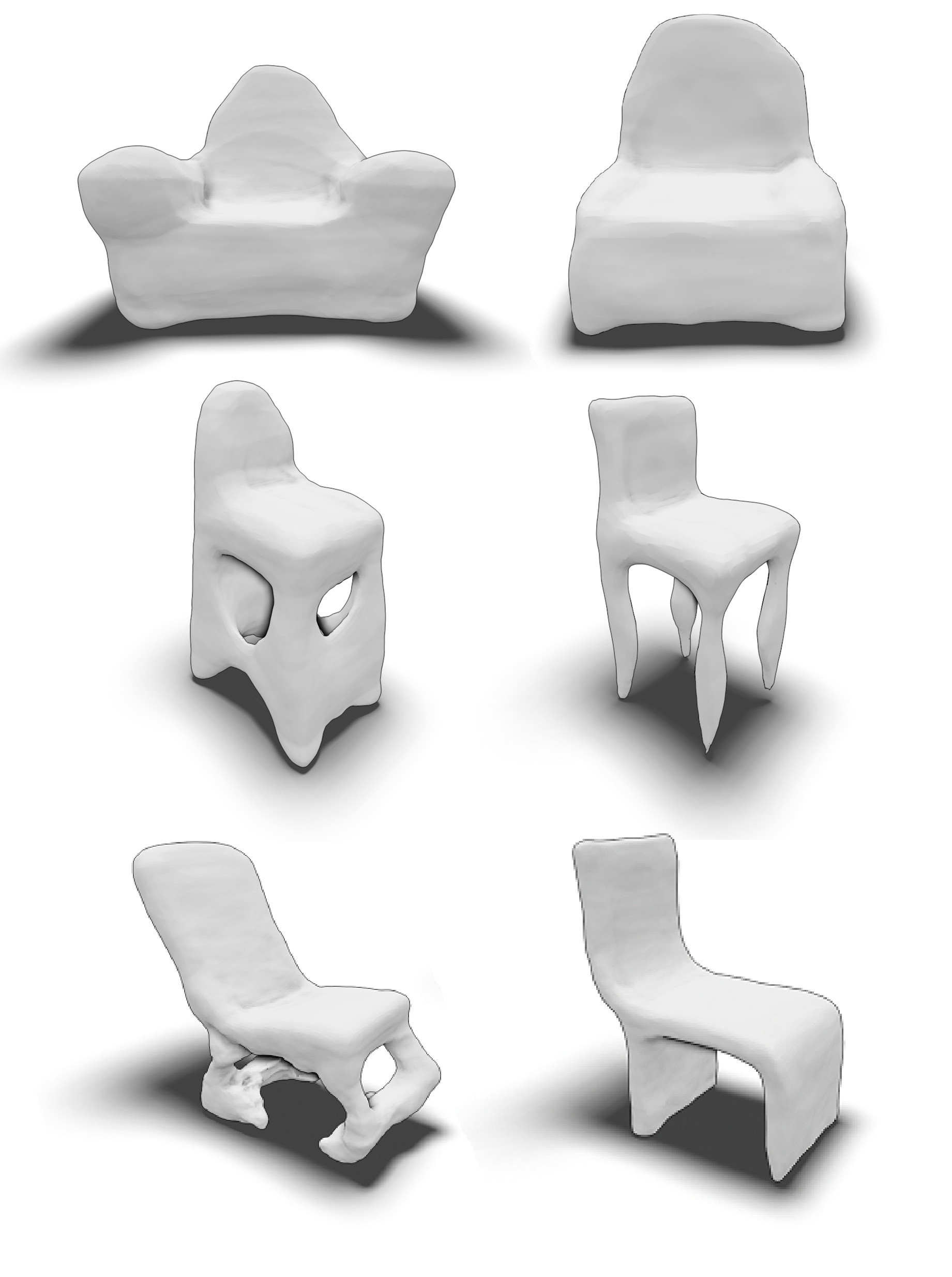}
        \caption{Generated shapes (left) and the closest item from the  shape collection.}\label{fig:most_crazy}
	\end{minipage}
\end{figure*}
\subsection{Comparison to Others} The main contribution of our work, the formulation of novel losses to obtain an editable, semantically meaningful latent space composed of 3D points and style, is difficult to assess with quantitative results alone. Hence, we compare to exemplary approaches from the state of the art regarding the method itself: 
DeepMetaHandles\cite{liu2021deepmetahandles} disentangle shapes into different parts that can be adapted individually (a \textit{meta-handle}). While they achieve good quality and their disentanglement into parts is meaningful, they do not offer direct control over the shape: Each meta-handle can be manipulated linearly, changing a part of the shape. This is different to directly controlling the shape surface, \ie actually moving the surface of the object into a target direction as our approach allows. Also, their disentanglement is not allowing style transfer as our work does. In summary, while they excel at outputting high qualitative shapes with only indirect control that is well suited for automated shape generation, our approach offers more direct editability of the surface, being better suited for \eg a 3D modelling tool.\\
Opposing to DeepMetaHandles, DeepMLS\cite{shechter2022deepmls} also offers direct manipulation of control points while only training on a single shape. In result, their editing can not infer certain semantics like symmetry of a airplane wings. As we train our network on a dataset, we do learn these semantics, and can also offer a much wider range of uses than just shape editing by also learning style - at the cost of lower quality outputs and requiring a whole dataset.\\
Lastly, Zhou \etal\cite{zhou2020unsupervised} produce a representation cleanly disentangled in pose and shape for deformable bodies through an autoencoder. Similar to our work, they also gain part of their disentanglement strength by requiring that mixed parts of latent code should be meaningful. While their results are more high-quality and their disentanglement works more consistent (\eg extreme poses are not problematic), they require multiple labeled examples of the same model.\\
In summary, it can be said that each of the discussed approaches, and generally most of the related work in the field, have their own use-cases and offer a well-working solution tailored to them. Compared to these other approaches, our method aims more at generality and versatility than quality, and excels at direct control and interpretability.

\subsection{Qualitative Analysis}
\begin{wrapfigure}[17]{r}{0.4\textwidth}
  \vspace{-45pt}
  \vspace{5pt}
  \centering
  \includegraphics[width=\linewidth]{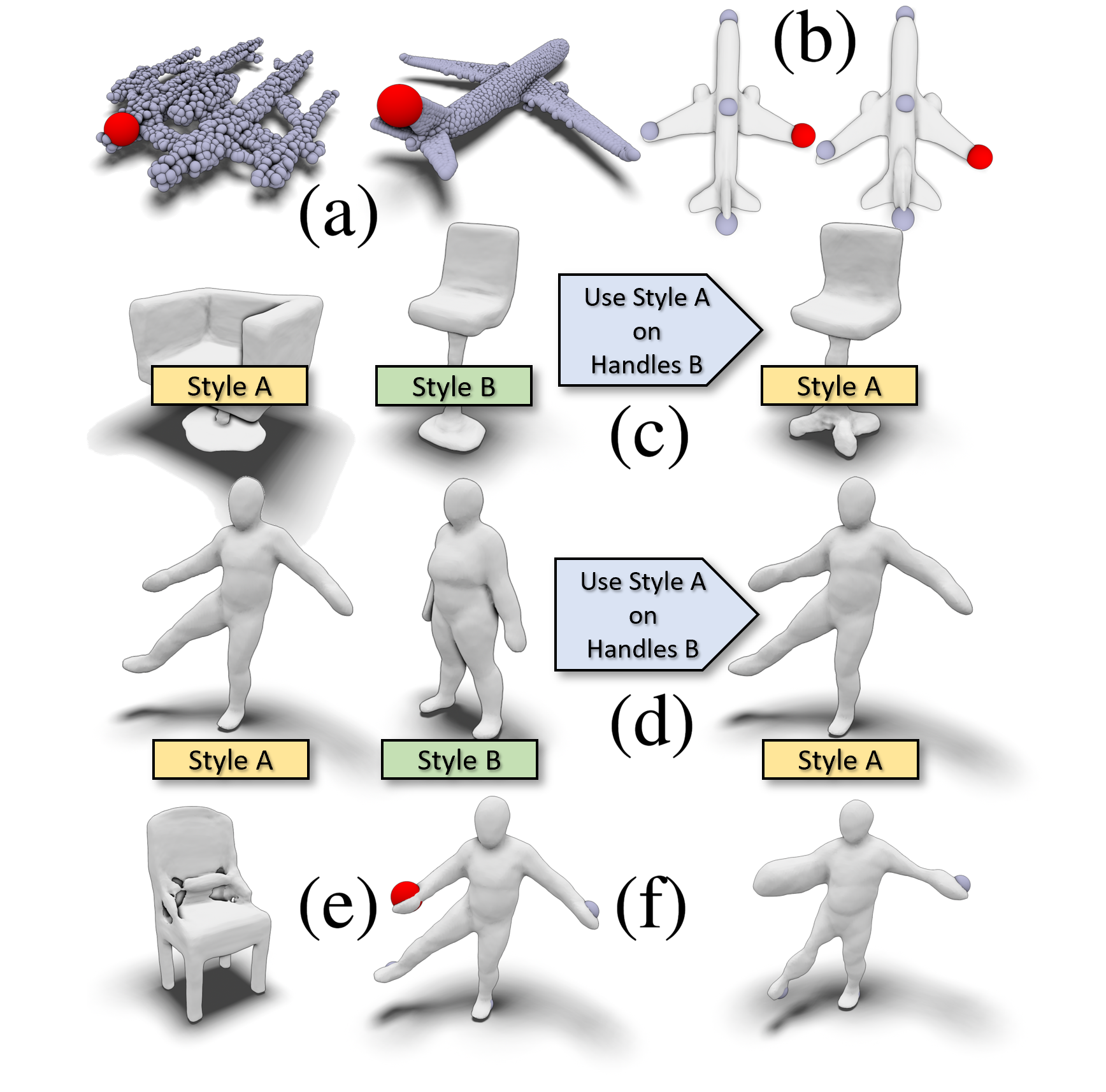}
  \caption{Examples of failure cases.}\label{fig:failures}
\end{wrapfigure}
We observe a large shape space that can produce many plausible variations that do not exist in the original shape collection, see \cref{fig:most_crazy}. For a larger collection of results, see Appendix \ref{app:more_results}.
Using the technique described in \cref{sec:canonical_handles}, we find that the produced handles are of mostly sufficient quality for man-made objects. While this does not hurt our generalization, for unique models, handles can placed inconsistent, see \cref{fig:failures}\textcolor{red}{a}. Especially organic shapes \eg from the FAUST Dataset, the handles show inconsistencies. Consequently, we hand-pick handles using the given correspondences for the FAUST Dataset.
\paragraph*{Disentanglement into style and handles}
The decomposition of an input shape into handles and style is strongly dependent on the number of handles used. Where \eg for a large number of handles on an airplane, the engines will have a handle controlling them instead of being encoded in the style.
Style transfer is also highly dependent on the setting it is used in and may not produce any useful characteristics. For chairs, the style is only meaningfully transferable for some examples (see \cref{fig:failures}\textcolor{red}{c}). For FAUST, style encodes mostly the body type, but breaks down for very unique poses like a flying kick (see \cref{fig:failures}\textcolor{red}{d}). Examples of style transfer can be seen in \cref{fig:styletransfer}. There exist shape collection given biases, \eg when changing the handle position of an airplane to be more bent backwards into delta wings, the engines will start to shrink and a rear engine can start to form (see \cref{fig:failures}\textcolor{red}{b}). A closer look at the shape collection shows that most airplanes with delta wings are having their engines tightly integrated into their wings, and airplanes with wings bent more backwards often have a rear engine. Hence, there is a plausible shift in the optics in what might be perceived as change in style.
\paragraph*{Structured analysis}
The reconstruction quality $L_{rec}$ together with the pre-trained handle encoder alone might already lead to good results. To show that this is not the case and our loss is compact, we provide the following ablation (see \cref{fig:ablation} for illustrations):
When not using a residual, a handle change may force unwanted change in the output to satisfy the proportionality of change in latent space versus change in output ($L_{d}$). This can be absorbed by the residual. Without $L_{ind}$, editing may fail because the net only plausibly decodes certain combinations of style and handles, and the manifold re-projection then just maps back to the original input. Using no style penalty, $L_{d}$ can be circumvented by \eg encoding everything in the style vector. In result, changes in handles can lead to no change in the output. Lastly, without $L_{d}$, we may get rather large changes in shape from small latent changes, \eg small handle shifts causing genus changes. We observe a significant increase in quality of the editing process when using all proposed losses.
\subsection{Quantitative Analysis}\label{sec:quant}
To quantify the abilities of our approach, we first validate the properties of the latent space we are training. We then demonstrate that our latent space is expressive in creating new shapes and furthermore is learning semantics by exploiting the learned latent space for segmentation.
\paragraph*{Latent Space Properties} To validate the desired/assumed properties of the latent manifold produced by our autoencoder and editing procedure, we provide two experiments on the chairs dataset of Shapenet\cite{chang2015shapenet}:\\
First, we show that our autoencoder has similar properties to a noisy autoencoder, \ie decoding and encoding a noised shape reduces the amount of noise. For this, we apply a random shift of distance in $[-0.03, 0.03]$ to each point, then de- and encode and measure the absolute mean difference to the input. For 1000 such operations on a random chair, we only have an average of $59.5$ percent and a median of $56.2$ percent of the noise remaining after a single re-projection. This shows properties similar to a noisy autoencoder, enabling us to re-project back to a plausible shape through encoding and decoding steps.\\
Second, we show that irregular/unique shapes are indeed worse for editing, \ie forming their own distinct region on the manifold of plausible shapes. While this does not hold for every unique shape, we can observe a general tendency towards "snapping back" to the old latent handle positions / latent in general when performing editing operations. This indicates a low editability. We demonstrate this by taking the $100$ examples from our shape collection where the handles are least/most unique, measured in distance to the next most similar handle constellation in euclidean distance. We then apply $10$ random deformations on each of the handle latent codes by applying a random shift in $[-0.1, 0.1]$, then de- and encode again. We measure that the unique chairs loses roughly $75$ percent of the edit applied through re-encoding, \ie mostly revert the edit applied, where as the least unique chairs only lose $58$ percent of the edit (for both average/median). Hence, a very unique chair is much less editable, having fewer plausible deformation directions in the latent space. This can be seen as an indication of a well-trained latent space: Single examples should generalize. In particular, \eg star destroyers as planes with low editability occurring in the dataset can be considered extreme cases (see \cref{fig:failures}\textcolor{red}{f} for an uneditable, unique pose in FAUST).
\paragraph*{Generative properties}
Liu \etal \cite{liu2021deepmetahandles} judge the resulting quality of their learned deformation space by splitting the shape collection in two groups $A$ (500 items) and $B$ (remaining items), where each item in $A$ is used to generate 20 variations, and $B$ is used to evaluate the quality of these variations. \textit{Coverage} (COV) describes how many different shapes from $B$ are the closest shape for any variation of $A$ under Chamfer distance, and \textit{minimum matching distance} (MMD)\cite{achlioptas2018learning} describes the minimum Chamfer distance of each item in $B$ to the closest item from the variations of $A$.\\
\begin{wraptable}{l}{90mm}
\centering
  \small
  \setlength{\tabcolsep}{5pt}
  \begin{tabular}{@{}l|lc|lc|lc@{}}
    \toprule
    Method & Chair & & Car & & Table\\
     & COV$\uparrow$ & MMD$\downarrow$ & COV$\uparrow$ & MMD$\downarrow$ & COV$\uparrow$ & MMD$\downarrow$\\
    \midrule
    3DN\cite{wang20193dn} & 32.0 & 4.56 & 46.6 & 2.91 & 30.6 & 4.26\\
    CC\cite{groueix2019unsupervised} & 51.0 & 4.26 & 50.3 & 2.79 & 50.2 & 3.88\\
    NC\cite{yifan2020neural} & 54.4 & 4.23 & 66.6 & 2.65 & 44.7 & 3.85\\
    DMH\cite{liu2021deepmetahandles} & 64.6 & 4.28 & 76.5 & 2.97 & 54.9 & 3.70\\
    Ours & 52.3 & 4.20 & 60.1 & 2.53 & 44.0 & 3.99\\
    \bottomrule
  \end{tabular}
  \caption{Results of computing COV and MMD on Shapenet Classes, COV in percent, MMD in Chamfer distance times 100. High COV and low MMD is desired. Other values taken from the work of Liu \etal \cite{liu2021deepmetahandles}.}
  \label{tab:metahandles}
\end{wraptable}
\begin{wraptable}{L}{90mm}
  \small
  \begin{tabular}{@{}lcccc@{}}
    \toprule
    Method & Airplane(3) & Chair(4) & Table(2) \\
    \midrule
    PointNet$\dagger$ \cite{qi2017pointnet} & 70.6 & 83.7 & 72.5\\
    PointNet++$\dagger$ \cite{qi2017pointnet++} & 72.2 & 86.1 & 74.1\\
    PointCNN$\dagger$ \cite{li2018pointcnn} & 71.7 & 83.4 & 62.5\\
    BAE-Net \cite{chen2019bae} & 61.1 & 83.7* & 78.7\\
    Ours & 67.5 & 65.7 & 73.8\\
    \bottomrule
  \end{tabular}
  \caption{Segmentation accuracy for (k) segments. $\dagger$ indicates using 5 percent of the data for labels, others are unsupervised. Numbers of the previous approaches taken from the work of Chen \etal \cite{chen2019bae}.}\label{tbl:bae}
\end{wraptable}
To generate a plausible variation, we gradually move one random handle towards the handle position from another random sample from $A$, using $4$ iterations while not holding the style vector fixated. See \cref{app:technical_cov_mmd} for technical details. What we lack in qualitative results compared to \cite{liu2021deepmetahandles}, we make up in usability, which is the main focus of our work. However, this is not truely measurable through MMD and coverage. For the (partially) better MMD values, we argue that the 3-dimensional degree of change we can apply to a handle can better fit certain shapes, as Liu \etal can only apply 1-dimensional changes.
\paragraph*{Segmentation}
We observed a close relationship between learned handles and semantic meaningful regions of the encoded shape. To use this relationship for unsupervised part segmentation, we apply a large number of small deformations through individual handles and measure the SDFs behavior at sampled surface points. These changes in the SDF then form feature vectors used to measure how different sample points are correlated, building a knn graph on the neighborhoods of points. On this graph, spectral clustering then yields a segmentation of the surface samples into a desired number of parts. Technical details can be found in Appendix \ref{app:segmentation}.
For a comparison of our unsupervised segmentation results, we use the shape collection and scores provided by BAE-Net\cite{chen2019bae}, see \cref{tbl:bae}. We compare in particular to BAE-Net, as it is also an unsupervised approach utilizing SDFs. They segment their input by having an autoencoder with multiple decoders, each one specializing on different segments of the shape collection.

Despite our approach solely operating on changes in the SDF, we obtain a semantically reasonable decomposition that is (in part) comparable to the state of the art. Examples can be seen in \cref{fig:segmentation_results}, which also shows a failure mode: As the ground truth is annotated by humans, we can obtain a segmentation that is different, but also plausible. Also note that the result of Chen \etal\cite{chen2019bae} on chairs is only possible as they join classes of chairs and tables during training. We did not perform such optimizations and expect further improvements to be possible.
We conclude that the active regions of individual handles are strongly related to high-level characteristics of the underlying shape domain.~\\

\section{Conclusion}
We introduce an autoencoder that learns to encode a manipulable latent representation from an input shape. Through tethering change in the latent space to change in the shape space, we decompose our latent representation into style and handles. We learn a representation that allows direct control for editing while implicitly learning semantics. Through projecting back to the learned data manifold, we can perform plausible editing within the range of the learned shape collection.
While the quality of the produced outputs is generally satisfying, we think that improvement through more powerful encoder and decoder architecture, more carefully chosen or augmented shape collections, and more consistent handle positions is possible.
We see possibilities to extend beyond the domain of geometry, \eg learning to manipulate photos of people by using facial landmarks as handles. Learning handles end-to-end instead of in a separate step could also increase performance and learned semantics.
The results show that the intuitive and disentangled latent space can be used for various tasks. This was demonstrated by performing various editing operations and style transfer on a qualitative, and segmentation and model generation on a quantitative level. We further provided quantitative evidence for the properties of the learned data manifold and the properties of our editing method. Lastly, we showed our unique direction of our work by comparing to other state of the art methods. We expect both improved quality and many more applications, \eg symmetry detection, to be possible.

%
%
\bibliographystyle{splncs04}
\bibliography{egbib}
\newpage
\appendix
\begin{figure}[h]
  \centering
   \includegraphics[width=0.8\linewidth]{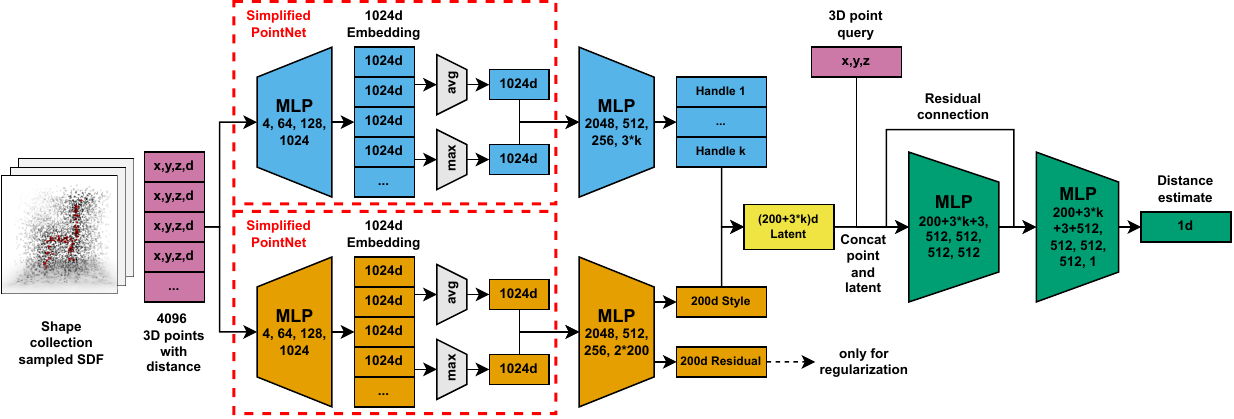}
    \caption{The detailed network architecture. Note that the style- and residual encoders (orange) share an embedding for memory reasons.}\label{fig:full_architecture}
\end{figure}
\begin{figure*}[h]
	\begin{minipage}{0.65\columnwidth}
		\centering
		\includegraphics[width=\textwidth]{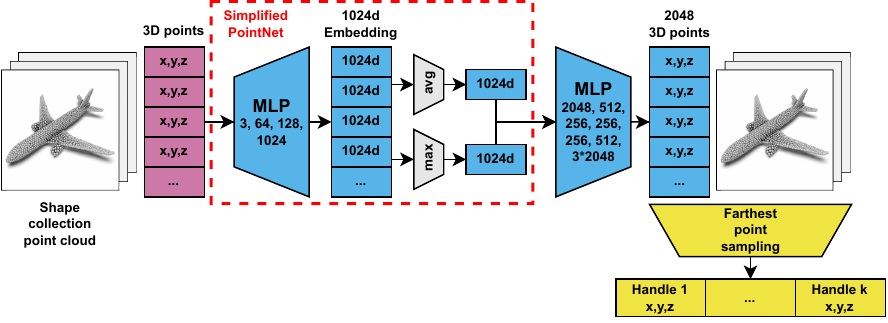}
		\caption{The simple architecture used to obtain correspondences between different shapes in a collection. The neurons of each output point specialize, with each point always being in a similar semantic part of the shape.}
		\label{fig:canonicaliser_a}
	\end{minipage}
	\hfill
	\begin{minipage}{0.3\columnwidth}
		\centering
		\includegraphics[width=\textwidth]{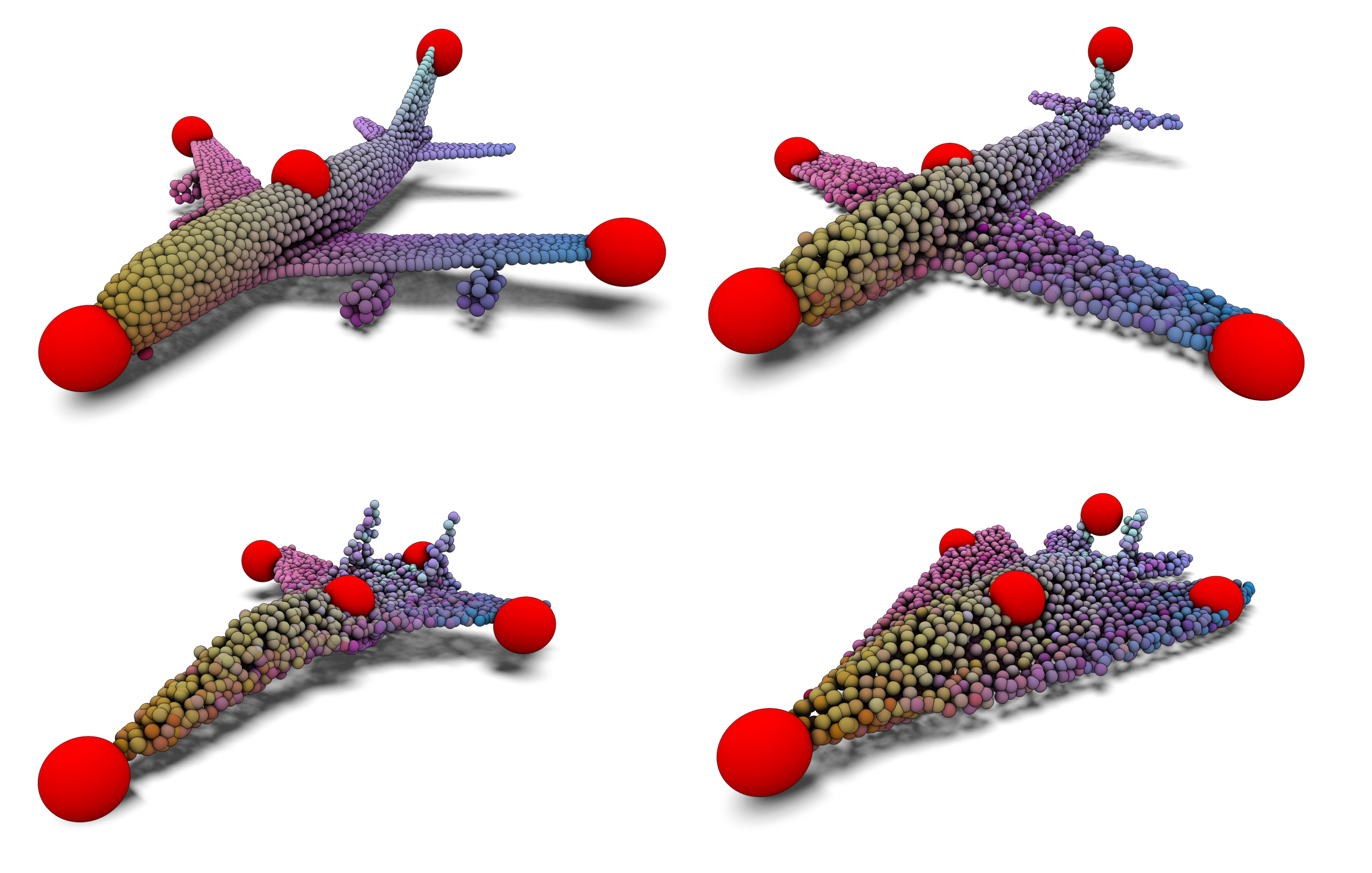}
		\caption{RGB coloring is computed on the top-left input per axis, mapping the relative X/Y/Z positions to RGB colors. Consistent colors then used across shapes.}\label{fig:canonicaliser_b}
		\label{label2}
	\end{minipage}
\end{figure*}
\section*{A. Implementation Details}\label{sec:details}\label{app:canonicaliser}
In this section, we give detailed information on how the samples are prepared, on the fine details of the architecture in \cref{fig:full_architecture}, on how our network is trained, and on how the output images seen throughout this work are extracted. We use the approach of Kleineberg \cite{Kleineberg2020} to sample our shape collections, as the Shapenet datasets are \eg not watertight. All shapes are centered and placed on the ground plane at $z=-1$ / airplanes at $z=-0.5$ while making sure the largest dimension of their bounding box is $2$ units wide, while taking SDF samples uniform randomly in the range of $[-1.1,1.1]$ for each coordinate axis.

\paragraph*{Architecture}
For the embedding, we use 1d convolution with kernel size 1 to increase from 3 (canonicalizer)/4 (main network) to (64, 128, 1024) channels per point, then applying both max pooling and average pooling to obtain two 1024 dimensional embedding vectors that we concatenate.
For the canonicalizer, this is then encoded to a 256-dimensional embedding vector through three linear layers (512, 256, 256), and then reconstructed to a full point cloud of 2048 3D points through 3 final linear layers (256, 512, 6144). The architecture and output can be seen in \cref{fig:canonicaliser_a}, a visualisation of the consistency can be found in \cref{fig:canonicaliser_b}. Note that the consistency is becoming weaker for unique shapes, see the rear of the bottom-right example. We would like to point that that this pre-training is just for increased performance, the whole network can be trained end-to-end (with the canonicalizer loss being added to the losses of the main network). However, this increases computation time significantly.\\
For the main network, we apply two separate encoders (one for handles, one joint for style and residual) on the embedding vector, reducing the number of channels to 512, then 256, then the desired target size. For all experiments in this paper, we used a 200 dimensional latent vector for style and residual each, while using 3 neurons per handle point. This can be altered without much impact on the quality, as long as the latent does not shrink below roughly 50 dimensions. For the decoder, we combine handles and style vector into a single latent vector, concatenate the desired point to decode, and then apply an MLP for 8 layers with channel size 512 and a residual connection from the latent vector to the 5th layer of the MLP. Note that this is the same decoder architecture as DeepSDF \cite{park2019deepsdf}. In total, our network had $4.5$ million parameters. For all layers, except the actual output layers and the layers that produce style, residual, and handles, we used a LeakyReLU \cite{zhang2017dilated} as non-linearity.\\
\paragraph*{Training and mesh extraction}
All networks are trained using the ADAM-W optimizer\cite{kingma2014adam}\cite{DBLP:journals/corr/abs-1711-05101} with AMSGrad\cite{reddi2019convergence}, weight decay rate of $0.005$ on the main network, batch size 16, no weight decay on the canonicalizer, learning rate $0.001$, tuned down to $0.0002$ after 500 epochs, and otherwise default parameters. We found that the networks were fairly robust regarding adaptations of the parameters and the varying domains of the shape collections used. No form of normalisation was used. All examples seen in this paper have been trained for 3 days on a single GeForce RTX 2080 Ti, resulting in between 1000 and 2300 epochs, depending on the shape collection. Training for longer periods of time did not show any significant improvements.\\
We extract all examples in this work by using marching cubes at a resolution of $200$ with a level of $0$ to $0.05$, depending on how filigree the output is. For example, thin parts like airplane wings rarely have a sample directly inside them. This results in the SDF at times never reaching to zero (or below). Using a level slightly larger than zero then yields a less fringed output.

\section*{B. Shape Segmentation Procedure}\label{app:segmentation}
\begin{figure}[t]
  \centering
   \includegraphics[width=0.8\linewidth]{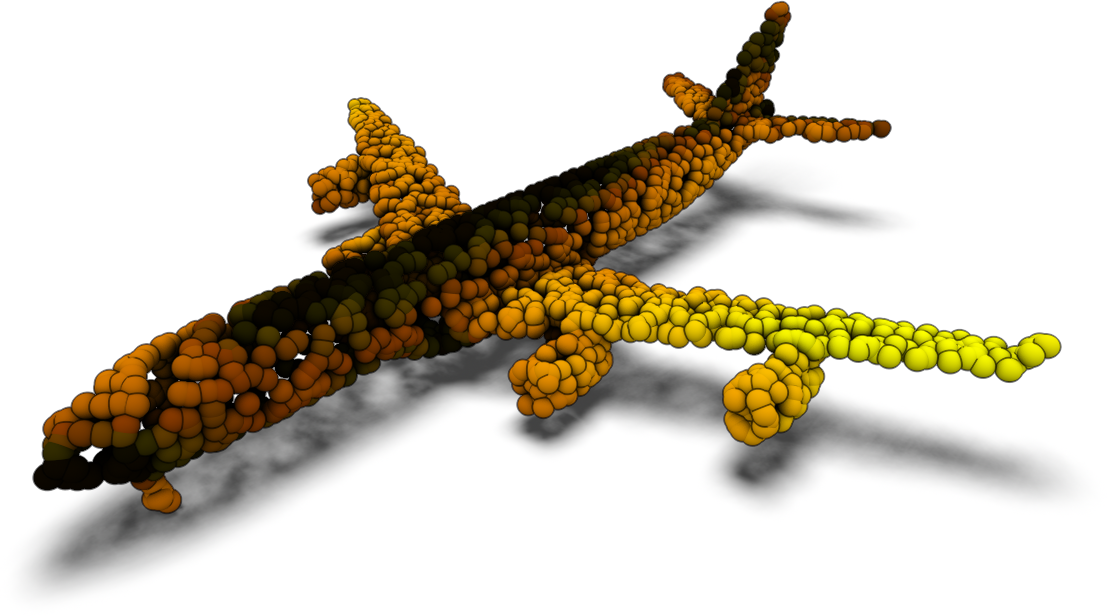}
    \caption{Response to 1024 random changes in one handle at the wing tip, here averaged. Brighter means more response.}\label{fig:response}
\end{figure}
To segment a given 3D mesh, we first need a distinct set of points where to measure the change of the SDF. In our comparison, we used the pre-computed sample points provided by Chen \etal \cite{chen2019bae} to better compare our results. Additionally, we sample the shapes SDF to compute the handle and style vector. 
Next we separately apply a Gaussian 3D shift to each handle and measure the resulting absolute change in the SDF value of each surface sample. To keep the shift magnitude in a reasonable range for each shape, at each segmentation we compute the standard deviation of all surface samples along the three axes and scale the shift by the minimum axis deviation value.
Repeating the random shifts and distance measurements $1024$ times yields a $h \times 1024$ feature descriptor per sample, where $h$ is the number of handles. 
Averaging this descriptor over the different shifts indicates the response of a samples distance value to a moving handle (see \cref{fig:response}). As some handles generally affect distance values more than others, but similar feature magnitudes are desired, all descriptor values originating from the same handle are divided by their joint $l1$-norm. To remove the large difference between highly active and almost inactive surface samples regarding distance changes, all values originating from the same sample are also divided by their joint $l1$-norm.
To obtain segmented parts, we build a knn-graph in the feature descriptor space with neighborhoods spanning over $5$ percent of the total number of surface samples. Each edge between neighboring samples $i$ and $j$ is given a similarity weight $s_{i,j}$ formulated as:
\begin{equation}
    s_{i,j} = e^{- \frac{\Vert f_i - f_j \Vert}{\Vert f_i \Vert \cdot \Vert f_j \Vert}}
\end{equation}
where $f_i$ and $f_j$ are the flattened feature descriptors of both samples.
Applying spectral clustering now yields an assignment for each of the points to one of the labels, which we in turn then can map back from the individual points to the input mesh.
In contrast to BAE-Net we do need to specify the exact number of expected parts instead of an upper limit for the chairs shape collection where some segments do not always occur (e.g.\ not all chairs have armrests). As spectral clustering does not output consistent indices for the part classes, we further apply a mapping to the ground truth indexing in the score calculation.

\section*{C. Technicalities of COV/MMD computation}\label{app:technical_cov_mmd}
The outputs are extracted by marching cubes and sampled with 4096 points, normalized to a unit sphere, and leveled to the ground plane. Liu \etal use $100000$ samples, resulting in computing $500 \times 20 \times |B|$ Chamfer distance values with a distance matrix each of $100000^2$ entries. To save resources, we only used $4096$ points. We observe that simply scaling the Chamfer distance with a coefficient computed on the shape collection yields the correct Chamfer distance within $~1/2/4$ percent error (table/chair/car) of the distance value computed with $100000$ samples. A derivation of Chamfer distance that takes the mean of euclidean distances instead is used.
\newpage
\section*{D. More results}\label{app:more_results}
\begin{figure*}
	\begin{minipage}{0.9\columnwidth}
		\centering
		\includegraphics[width=\textwidth]{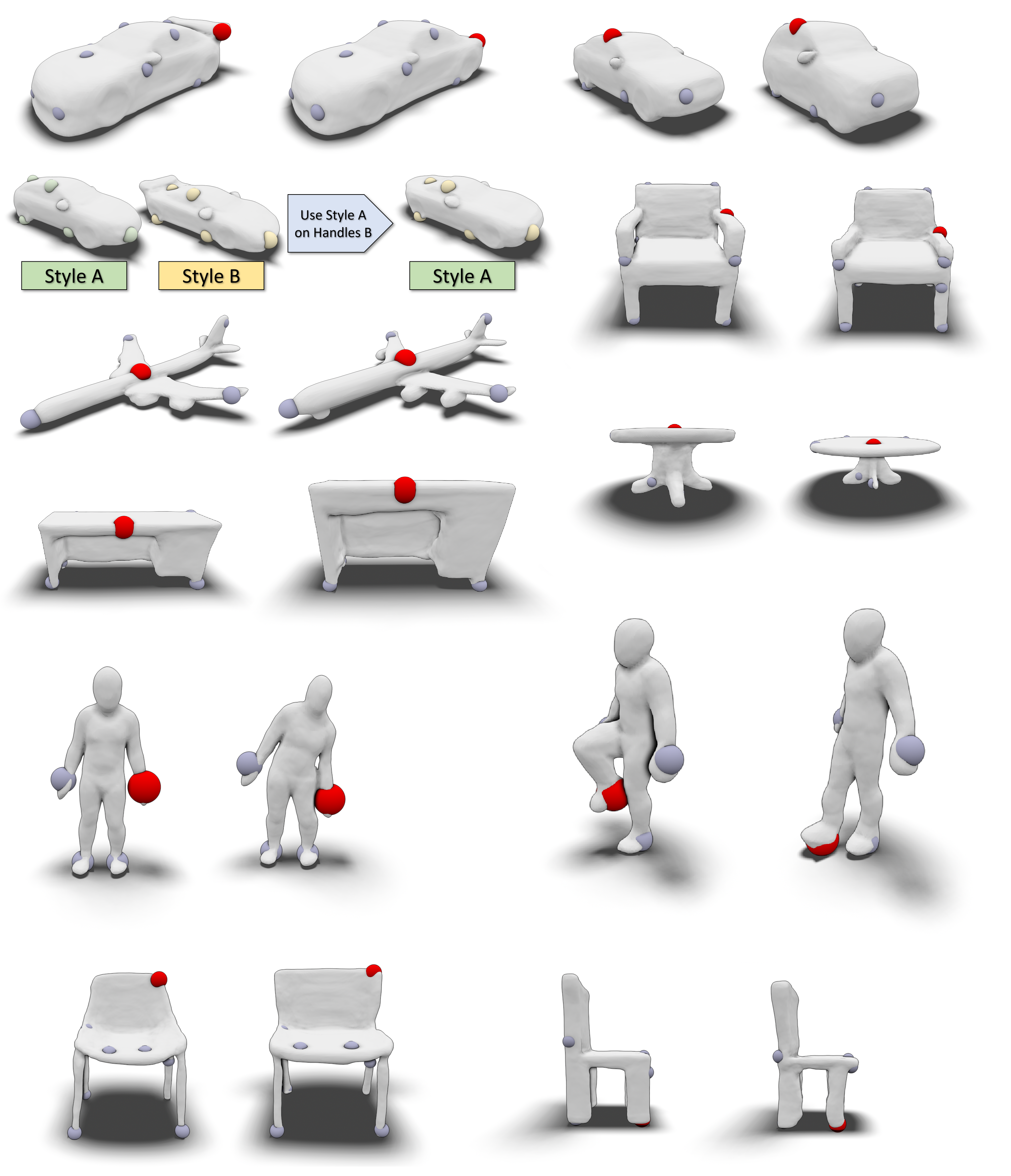}
		\caption{More selected results from the editing process.}\label{fig:extended_results}
	\end{minipage}
\end{figure*}

We show more results from our editing procedure in \cref{fig:extended_results}. Note how the car on the top-left is using 16 handles, showing the spoiler being part of the handle configuration. Using 8 handles, the spoiler instead becomes part of the style, as can be seen below.

\end{document}